\documentclass[11pt]{article}

\usepackage{natbib}
\usepackage[margin=1in]{geometry}
\usepackage{amsmath,amssymb,latexsym}
\usepackage{graphicx}
\usepackage{multirow}
\usepackage{booktabs} 
\usepackage{hyperref}

\newcommand{\bbeta}{\ensuremath{\boldsymbol{\beta}}}
\newcommand{\btheta}{\ensuremath{\boldsymbol{\theta}}}
\newcommand{\bxi}{\ensuremath{\boldsymbol{\xi}}}
\newcommand{\bXi}{\ensuremath{\boldsymbol{\Xi}}}
\newcommand{\bmu}{\ensuremath{\boldsymbol{\mu}}}

\newcommand{\bleta}{\ensuremath{\boldsymbol{\eta}}}

\title{Equity-Directed Bootstrapping: Examples and Analysis}
\author{Harish~S. Bhat\thanks{Applied Mathematics, University of California Merced, Merced, CA 95343} \footnote{email: hbhat@ucmerced.edu} \and Majerle~E. Reeves$^\ast$ \and Sidra Goldman-Mellor\thanks{Public Health, University of California Merced, Merced, CA 95343}}

\begin{document}
\maketitle

\begin{abstract}
When faced with severely imbalanced binary classification problems, we often train models on bootstrapped data in which the number of instances of each class occur in a more favorable ratio, {\it e.g.}, one.  We view algorithmic inequity through the lens of imbalanced classification: in order to balance the performance of a classifier across groups, we can bootstrap to achieve training sets that are balanced with respect to both labels and group identity.  For an example problem with severe class imbalance---prediction of suicide death from administrative patient records---we illustrate how an equity-directed bootstrap can bring test set sensitivities and specificities much closer to satisfying the equal odds criterion.  In the context of na\"ive Bayes and logistic regression, we analyze the equity-directed bootstrap, demonstrating that it works by bringing odds ratios close to one, and linking it to methods involving intercept adjustment, thresholding, and weighting.
\end{abstract}

\section{Introduction}
\label{sect:intro}
Many real-world predictive modeling problems feature class imbalance, {\it e.g.}, prediction of rare diseases from health care records \citep{Schubach2017}, biological activity of pharmaceutical compounds \citep{Esposito2021}, and clicks on online advertising \citep{McMahan2013}.  Researchers have developed a variety of methods to deal with class imbalance, employing techniques such as weighting, bootstrapping, or threshold adjustment  \citep[Chap. 16]{Kuhn2018}.  In past work, many authors have examined the empirical performance of such strategies in conjunction with popular  classification methods, comparing them on both synthetic and real data sets \citep{chawla2002smote, tang2008svms, krawczyk2014cost, Branco2016, zou2016finding, Hasanin2019}.  For methods that admit a compact mathematical description, such as logistic regression and linear discriminant analysis, we also find theoretical results for imbalanced problems \citep{KingZeng2001, Xue2015, Wang2020}.

We focus on a problem with a clear class imbalance, predicting death by suicide from patients' emergency room visits.  In our overall database, more than $99.9\%$ of the data corresponds to patients who do not die by suicide (the negative class, $Y=0$).  A more subtle problem we encounter relates to algorithmic fairness \citep{Mehrabi2021}.  If we address the class imbalance (between $Y=1$ and $Y=0$ labels) but do not address racial/ethnic subgroup-level differences, the trained models show algorithmic bias, measurable using the equal odds criterion \citep{Hardt2016}.  If, hypothetically, a prediction of high suicide risk for a patient results in opportunities to receive information and interventions (such as a post-discharge phone call), then these opportunities will be allocated in an unequal way.

We seek models for which the opportunity to receive treatment is independent of racial/ethnic group identity, conditional on the true outcome.  In this idealized setting, all groups would have equal opportunities to receive outreach services and interventions to prevent suicide death.  As a concrete step towards this ideal, we propose an equity-directed bootstrapping procedure that aims to balance both test set sensitivity and specificity across groups, aiming to satisfy the equal odds criterion \citep{Hardt2016}.  The philosophy behind this procedure is to view algorithmic inequity as a type of imbalance that goes beyond the more typically analyzed class imbalance.  The equity-directed bootstrap generates training sets that are balanced with respect to class label \emph{and} racial/ethnic group membership.  We will see that in practice, for a variety of predictive modeling approaches, this procedure results in test set sensitivities and specificities that vary less between subgroups.

In order to further motivate the problem, we begin in Section \ref{sect:data} by presenting a data set, fairness metrics, and comparison of four preliminary models.  In Section \ref{sect:equityboot}, we describe the equity-directed bootstrap.  This procedure is designed to improve algorithmic equity for any predictive modeling technique; we verify this for the four models considered earlier.  In Section \ref{sect:theory}, we analyze the equity-directed bootstrap, in the specific cases where the bootstrap is paired with either the na\"ive Bayes or logistic regression classifiers.  We offer mathematical and computational explanations of how and why the equity-directed bootstrap works in general: in short, the bootstrap leads to logistic regression models whose odds ratios are nearly one across all groups.  We relate the bootstrap to methods from the imbalanced classification literature: intercept adjustment, thresholding, and weighting.  We conclude in Section \ref{sect:discussion}.

\section{Data, Metrics and Preliminary Modeling}
\label{sect:data}

\begin{table}[t]
\begin{center}
\begin{tabular}{lrrr}
               &    APR &   Suicide Deaths &   Suicide Deaths / 100K      \\
\hline
 White         & 22792782 &   30365 & 133  \\
 Hispanic      & 12681054 &    5175 &  41  \\
 Black         &  5638578 &    2225 &  39  \\
 Asian         &  2570990 &    1334 &  52  \\
 Native American         &   161502 &   275 & 170  \\
\hline
\end{tabular}
\caption{ Data broken down by race/ethnic feature, excluding the \emph{Other} and \emph{unknown} race categories. Suicide rates differ by group.}
\label{race_break}
\end{center}
\end{table}

This study uses deidentified administrative patient records (APRs) provided by the California Office of Statewide Health Planning and Development (OSHPD) together with linked death records provided by the California Department of Public Health (CDPH) Vital Records. This study was approved by Institutional Review Boards of the California Health and Human Services Agency and the University of California, Merced.  We analyze all visits to all California-licensed emergency departments (EDs) from 2009-2013, by individuals aged at least 5 with a California residential zip code. The data contains N = 44,872,599 records from 14,716,914 patients, and includes the date and underlying cause of death for all decedents who died in California in 2009-2013.

To the $i$-th record (for $i = 1, \ldots, N$), we assign a label of $y_i=1$ if the record corresponds to a patient who died by suicide during the period 2009-2013; otherwise, we assign a label of $y_i=0$.  We determine whether a patient has died by suicide by 
checking the cause of death for ICD-10 codes X60-X84, Y87.0, or U03.  Let $\mathbf{y}$ be the vector whose $i$-th entry is $y_i$.

In Table \ref{tab:predictors}, we detail the predictors we extracted from the raw data.  
For categorical variables, we employ one-hot or dummy encoding.  Taking this into account, the total dimension of the predictor space is $612$.  Let $\mathbf{X}$ denote the $N \times 612$ matrix of all predictors.  Though some patients make multiple visits in our data set, corresponding to multiple rows of $(\mathbf{X},\mathbf{y})$, we treat each row $(\mathbf{x}_i,y_i)$ as an independent sample from random variables $(X,Y)$.

The APRs include up to $25$ Clinical Classifications Software (CCS) diagnostic codes. These CCS codes aggregate more than 14,000 International Classification of Diseases-version 9 (ICD-9-CM) diagnoses into 285 mutually exclusive and interpretable category codes, only 262 of which appear in our data.   Each visit also includes up to $5$ E-Codes, which provide information about the intent (accidental, intentional, assault, or undetermined) of external injuries and poisonings.  The APRs omit information such as vital signs and height/weight found in full medical records.

For each visit, the diagnostic (respectively, E-Code) predictors consist of the \emph{union} (or logical \texttt{OR}) of the one-hot encodings of the $\leq 25$ diagnostic codes (respectively, $\leq 5$ E-Codes) assigned to that visit.  We supplement the raw variables with $20$ domain expert-provided features, one of which is the numeric/continuous Charlson comorbidity index \citep{Charlson1987}, and $19$ of which are binary variables that record the presence of various conditions, {\it e.g.}, suicidal ideation, self-harm, HIV/AIDS, congestive heart failure, etc.

Our goal is to use $(\mathbf{X}, \mathbf{y})$ to train models that estimate the conditional probability $P(Y=1 \, | \, X = x)$.  In this task, we envision that the model uses information from a single patient visit ($X=x$) to compute a probability of offering treatment or intervention to that patient.  In our records, 9,736 patients (corresponding to 40,831 records) have died by suicide; as $<0.1\%$ of the data is in the $Y=1$ (death by suicide) class, the classification problem is imbalanced.   We break these patient records down by sensitive racial/ethnic attribute in Table \ref{race_break}. Because there are not enough samples of the positive class to make generalizable predictions, we omit the Native American group from the analysis below.  

 \setlength{\tabcolsep}{2pt}

\begin{table}[t]
\begin{center} \small
  \begin{tabular}{ll|ll}
    \textbf{description} & \textbf{type} & \textbf{description} & \textbf{type} \\ \midrule
    age & numeric & insurance category & categorical (5 levels) \\
    ED visit & binary & disposition (ED) & categorical (34 levels) \\
    PD visit & binary & disposition (PD) & categorical (14 levels) \\
    facility ID number & numeric & payer (ED) & categorical (22 levels) \\
    facility ZIP & numeric & facility county (ED) & categorical (58 levels) \\
    corrected ZIP & numeric & type of care & categorical (6 levels) \\
    hospital ZIP & numeric & source site & categorical (10 levels) \\
    rural/urban score & numeric & admission type & categorical (5 levels) \\
    length of stay & numeric & payer category & categorical (10 levels) \\
    pay plan & numeric & payer type & categorical (4 levels) \\
    domain-expert features (20) & one numeric and $19$ binary & patient county & categorical (58 levels) \\
    present on arrival & binary & hospital county & categorical (58 levels) \\
    sex & categorical (4 levels) & CCS diagnostic codes & categorical (262 levels) \\
    race & categorical (7 levels) & E-codes &  categorical (24 levels) \\
  \end{tabular}
  \caption{Each row of our data set corresponds to a unique patient visit; for each visit, our data includes these predictors.  The total dimension of the predictor space, accounting for one-hot encoding of all categorical variables, is $612$.  Here ED stands for emergency department while PD stands for patient discharge and in-patient hospitalization.}
  \label{tab:predictors}
    \end{center}
\end{table}

\subsection{Fairness Metrics}
\label{sect:metrics}
As we build predictive models for suicide death ($Y$) using various $N \times p$ training matrices (each with $p=612$), we seek to quantify the extent to which our models satisfy established criteria for algorithmic fairness.  We now review one such set of criteria, in the context of our data and predictive task.  Let $A \in \mathcal{A}$ denote racial/ethnic identity (or, more generally, membership in any subgroup), let $\hat{Y}$ denote the predictions of a trained model, and let $Y$ denote (as above) the true label.  Treating $A$, $\hat{Y}$, and $Y$ as discrete random variables, we express our idealized goal as
\begin{equation}
\label{eqn:independent}
P(\hat{Y} = 1 \, | \, A=a, Y=y) = P( \hat{Y} = 1 \, | \, Y=y) \text{ for } y \in \{0, 1\}.
\end{equation}
If, hypothetically, $\hat{Y}=1$ leads to offers of treatment for a patient, then (\ref{eqn:independent}) says that, \emph{conditional on the patient's true outcome, the probability of being offered treatment is independent of the patient's racial/ethnic identity}.  As it is difficult to quantify independence of random variables in the finite-sample setting, we work instead with the \emph{equal odds} criterion \citep{Hardt2016}: for $y \in \{0,1\}$ and all $a_1, a_2 \in \mathcal{A}$,
\begin{equation}
\label{eqn:equalodds}
P(\hat{Y} = 1 \, | \, A=a_1, Y=y) = 
P(\hat{Y} = 1 \, | \, A=a_2, Y=y).
\end{equation}
Note that (\ref{eqn:independent}) implies (\ref{eqn:equalodds}).
If we enforce (\ref{eqn:equalodds}) only for instances such that $Y=1$ (patients who have died by suicide), we obtain the weaker \emph{equal opportunity criterion}: for all values of $a_1$ and $a_2$,
\begin{equation}
\label{eqn:equalopp}
P(\hat{Y} = 1 \, | \, A=a_1, Y=1) = 
P(\hat{Y} = 1 \, | \, A=a_2, Y=1).
\end{equation}
The probabilities in (\ref{eqn:independent}-\ref{eqn:equalopp}) are all with respect to conditional distributions of $\hat{Y}$.  If we approximate the probabilities in (\ref{eqn:equalodds}) empirically---with $\hat{y}_i$, $a_i$, and $y_i$ denoting, respectively, the predicted label, group identity, and true label for the $i$-th instance of a data set with $N$ instances---then we obtain
\begin{equation}
\label{eqn:empiricalequalodds}
\frac{\sum_{i=1}^N I_{\hat{y}_i=1} I_{a_i=a_1} I_{y_i=y}}{\sum_{i=1}^N I_{a_i=a_1} I_{y_i=y}} = \frac{\sum_{i=1}^N I_{\hat{y}_i=1} I_{a_i=a_2} I_{y_i=y}}{\sum_{i=1}^N I_{a_i=a_2} I_{y_i=y}} \text{ for } y \in \{0,1\} \text{ and all } a_1, a_2 \in \mathcal{A}.
\end{equation}
We see that (\ref{eqn:empiricalequalodds}) is equivalent to balancing both sensitivity (for $y=1$) and specificity (for $y=0$) across all elements of $\mathcal{A}$, i.e., across all racial/ethnic groups.  Strictly speaking, evaluating (\ref{eqn:empiricalequalodds}) for $y=0$ yields an equality of false positive rates (FPRs), but note that specificity is $1-\text{FPR}$.  Based on this, for each $a \in \mathcal{A}$, let
\begin{equation}
\label{eqn:sensspeca}
\operatorname{sens}_a = \frac{\sum_{i=1}^N I_{\hat{y}_i=1} I_{a_i=a} I_{y_i=1}}{\sum_{i=1}^N I_{a_i=a} I_{y_i=1}}, \text{ and } \operatorname{spec}_a = 1 - \frac{\sum_{i=1}^N I_{\hat{y}_i=1} I_{a_i=a} I_{y_i=0}}{\sum_{i=1}^N I_{a_i=a} I_{y_i=0}}.
\end{equation}
Then we can quantify the degree to which a given model violates (\ref{eqn:empiricalequalodds}) by computing any measure of the dispersion of that model's $\operatorname{sens}_a$ and $\operatorname{spec}_a$ over all $a \in \mathcal{A}$.  In this paper, we use range to measure this dispersion.  For a given model, if the ranges of $\operatorname{sens}_a$ and $\operatorname{spec}_a$ both vanish, then that model satisfies (\ref{eqn:empiricalequalodds}), {\it i.e.}, the empirical version of the equal odds criterion.

\begin{table}[t]
\begin{center}
\begin{tabular}{lrrrrr|rrrrr}
& \multicolumn{5}{c}{\bfseries Specificity} & \multicolumn{5}{c}{\bfseries Sensitivity} \\
 & Black & Asian & White & Hispanic & Range & Black & Asian & White & Hispanic & Range \\ \hline
Logistic Regression & 0.83 & 0.77 & 0.31 & 0.83 & 0.52 & 0.57 & 0.68 & 0.91 & 0.48 & 0.43 \\
Na\"ive Bayes & 0.78 & 0.76 & \textbf{0.35} & 0.78 & \textbf{0.43} & \textbf{0.61} & 0.63 & 0.88 & 0.54 & \textbf{0.34} \\
XGBoost & 0.84 & 0.75 & 0.33 & 0.79 & 0.51 & 0.58 & \textbf{0.72} & 0.93 & \textbf{0.56} & 0.37 \\
Random Forest & \textbf{0.94} & \textbf{0.97} & 0.24 & \textbf{0.94} & 0.73  & 0.29 & 0.28 & \textbf{0.96} & 0.21 & 0.75 \\ 
\end{tabular}
\caption{Using a blind training set $B$ that addresses class imbalance but ignores racial/ethnic group membership, we train four predictive models.  After adjusting model-specific thresholds $\tau$ such that training set specificities are nearly equal to $0.56$, and after adjusting hyperparameters so as to maximize performance on validation sets, we compute group-specific sensitivities and specificities using the positive and negative test sets $T_a^j$.  For each model, we compute the range (max minus min value) of the specificity and sensitivity across the four racial/ethnic groups.  The larger the range, the greater the model's violation of the empirical equal odds criterion (\ref{eqn:empiricalequalodds}).  Boldface indicates the best result in each column. \emph{Our chief finding is that ignoring racial/ethnic group membership results in models with much higher sensitivity for White patients, and much lower sensitivities for patients in the Black, Asian, and Hispanic groups.}}
\label{tab:blind}
\end{center}
\end{table}

\subsection{Preliminary Modeling}
\label{sect:prelimmod}
We first split the entire $(\mathbf{X},\mathbf{y})$ data set by racial/ethnic group $a \in \mathcal{A} = \{1,2,3,4\}$ and positive/negative label $y \in \{0,1\}$, resulting in $8$ subsets $\{G_a^j\}$.  Within each subset $G_a^j$, we employ a 60/20/20 sequential split into training, test, and validation sets $R_a^j$, $T_a^j$, and $V_a^j$.  The test and validation sets are frozen.  Because they have already been split by group $a$, the $T_a^j$ sets can easily be used to compute the test set $\operatorname{sens}_a$ and $\operatorname{spec}_a$ from (\ref{eqn:sensspeca}).

As a preliminary model, suppose we ignore racial/ethnic group membership and focus exclusively on class imbalance.  One way to do this is to form positive and negative training sets $R^j = \cup_{a} R_a^j$.  Here $R^1$ has $\approx24498$ rows, while $R^0$ has $\approx26899061$ rows.  To form a balanced training set, we sample $1$ million rows (with replacement) from $R^1$ and $1$ million rows (without replacement) from $R^0$.  Let $B$ denote the resulting training set (with $2$ million total rows).  We call this the blind bootstrapping approach.

With the class-balanced training set $B$, we train four predictive models, in turn: logistic regression, na\"ive Bayes \citep[Chap. 6]{ESL2}, extreme gradient boosting or XGBoost \citep{Chen2016}, and random forests \citep{Breiman2001}.  As each of these models outputs an estimated probability $\hat{P}$ of membership in the positive class, we introduce a threshold $\tau$; then the predicted class is $\hat{Y} = 1$ if $\hat{P} \geq \tau$, and $\hat{Y} = 0$ if $\hat{P} < \tau$.

For each model, we adjust $\tau$ so that the training set \emph{specificities} are as close to equal as possible; in this study, we choose $0.56$ as our target training set specificity.  This approximately equalizes average \emph{test} set specificities, enabling us to compare models on the basis of test set sensitivity.  We also choose hyperparameters that maximize performance on the validation sets $V_a^j$.  Once hyperparameter choices have been finalized, we evaluate the sensitivities and specificities of each trained model on the \emph{test sets} $T_a^j$.

We report the test set findings of this preliminary modeling in Table \ref{tab:blind}.  Our main finding is that all of the models fail to satisfy the empirical equal odds criterion (\ref{eqn:empiricalequalodds}).  When training with the Blind method, White patient files (a majority of the data set) are much more likely to be classified as positive for suicide death regardless of the true label (high sensitivity, low specificity). The opposite is true for the Hispanic, Black, and Asian groups, with patient files much less likely to be positive for suicide death regardless of true label (low sensitivity, high specificity). This model's overreliance on race/ethnicity features dominates whatever it learns about other features that predict suicide death.

The construction of the blind training set $B$ combines downsampling of the majority class ($R^0$) with upsampling of the minority class ($R^1$).  We have experimented with a pure upsampling approach that includes all records in the majority class, resulting in a training set that is roughly $26$ times larger than $B$.  This increased training set size greatly increases the computational time required to train each model, without improving results much.  In what follows, we avoid pure upsampling approaches for this reason.

Our overall findings in this section match those of \cite{YatesColey2021}; note that their data set is restricted to patients making mental health visits, while our data includes visits for any reason.  Still, \cite{YatesColey2021} find high sensitivity for random forest and logistic regression (with LASSO) models for White, Hispanic, and Asian patients; the same models yield poor sensitivity for Black patients, American Indian/Alaskan Native patients, and patients whose racial/ethnic information is missing.  Taken together, these findings motivate us to develop a procedure that reduces the algorithmic bias of predictive models for suicide death.  We describe this procedure next.

\begin{table}
\begin{center}
\begin{tabular}{lrrrrr|rrrrr}
& \multicolumn{5}{c}{\bfseries Specificity} & \multicolumn{5}{c}{\bfseries Sensitivity} \\
 & Black & Asian & White & Hispanic & Range & Black & Asian & White & Hispanic & Range \\ \hline
Logistic Regression & \textbf{0.57} & 0.61 & 0.51 & 0.56 & 0.10 & 0.77 & 0.77 & 0.79 & \textbf{0.78} & \textbf{0.02} \\
Na\"ive Bayes & 0.55 & \textbf{0.64} & \textbf{0.52} & 0.54 & 0.12 & \textbf{0.82} & 0.71 & 0.77 & 0.76 & 0.11 \\
XGBoost & 0.56 & 0.60 & \textbf{0.52} & 0.57 & 0.08 & 0.77 & \textbf{0.81} & \textbf{0.82} & \textbf{0.78} & 0.05 \\
Random Forest & \textbf{0.57} & 0.57 & \textbf{0.52} & \textbf{0.58} & \textbf{0.06}  & 0.72 & 0.79 & 0.77 & 0.74 & 0.07 \\ 
\end{tabular}
\caption{Using the equity-directed bootstrap training set $E$ that addresses class imbalance \emph{and} racial/ethnic group membership, we train four predictive models.  After adjusting model-specific thresholds $\tau$ such that training set specificities are nearly equal to $0.56$, and after adjusting hyperparameters so as to maximize performance on validation sets, we compute group-specific sensitivities and specificities using the positive and negative test sets $T_a^j$.  For each model, we compute the range (max minus min value) of the specificity and sensitivity across the four racial/ethnic groups.  Boldface indicates the best result in each column. \emph{Our chief finding is that the equity-directed bootstrap yields models that are much closer to satisfying the empirical equal odds criterion (\ref{eqn:empiricalequalodds}) than the models built with the blind training set $B$.}}
\label{tab:equity}
\end{center}
\end{table}

\section{Equity-Directed Bootstrap: Method and Results}
\label{sect:equityboot}
We can generalize the construction described at the beginning of Section \ref{sect:prelimmod}.  Given the entire $(\mathbf{X},\mathbf{y})$ data set, we first split the data set by both $a \in \mathcal{A}$ and by positive/negative label $y \in \{0,1\}$.  Let $|\mathcal{A}|$ denote the number of elements in $\mathcal{A}$.  Then the result of this first split is $2 |\mathcal{A}|$ subsets $G_a^j$.  We further split each $G_a^j$ into training, test and validation sets $R_a^j$, $T_a^j$, and $V_a^j$.

With this setup, we can easily describe the equity-directed bootstrap: \textbf{sample $M$ rows from each $R_a^j$}. If there are at least $M$ rows in a given $R_a^j$, we have the option of sampling either with or without replacement; otherwise, we sample with replacement. \textbf{The union of the resulting samples is the equity-directed bootstrap training set $E$}.  Note that $E$ is balanced across racial/ethnic groups and across positive/negative labels.

Using the class- and group-balanced training set $E$, we retrain our  four predictive models.  As before, we develop model-specific thresholds $\tau$ to equalize training set specificities at $0.56$, and we use the validation set to optimize hyperparameters for each model.  Once hyperparameter choices have been finalized, we evaluate the sensitivities and specificities of each trained model on the \emph{test sets} $T_a^j$.

In Table \ref{tab:equity}, we report our test set findings.  Note that for each predictive modeling technique, the specificities and sensitivities vary little as a function of racial/ethnic identity.  The values of the Range have decreased by an average of 83\% from Table \ref{tab:blind} to Table \ref{tab:equity}.  While the equity-directed bootstrap does not yield test set Range values that are exactly zero, it does move the models into a regime where they nearly satisfy the empirical equal odds criterion (\ref{eqn:empiricalequalodds}).

In this study, we prioritize sensitivity for two reasons.  First, it is simple to create a model with high accuracy and perfect specificity at the expense of zero sensitivity---simply predict that no patients die by suicide.  The more challenging and interesting task is to develop a model with high sensitivity without overly sacrificing specificity.  Second, we envision that possible interventions would be non-intrusive ({\it e.g.}, post-discharge phone calls or postcards) and that the penalty for a false negative---missing a true death by suicide---would greatly exceed the penalty for a false positive---offering the intervention to those who are not at high risk for death by suicide.

With this in mind, we notice from Table \ref{tab:equity} that the test set sensitivity for the White population has declined from values in the interval $[0.88, 0.96]$ in Table \ref{tab:blind} to values in the interval $[0.79, 0.82]$ in Table \ref{tab:equity}.   This decrease is consistent with theoretical results.  Suppose that we have a trained model that achieves demographic parity on our data set, {\it i.e.},  the predictions $\hat{Y}$ are independent of racial/ethnic identity $A$.  Then the model's joint error across racial/ethnic groups is bounded below \citep{Zhao2019}---in short, one can achieve demographic parity but only by trading off some accuracy. We hypothesize that a similar tradeoff holds for equalized odds, and that some decline in sensitivity for the White population  may be necessary to achieve empirical equalized odds.

Examining the logistic regression model coefficients for racial/ethnic predictors (Table \ref{tab:race_coeff}), we see that with the Blind training set B the logistic regression model learns to predict future suicide death for White patients at a much higher rate than any other racial/ethnic group. Alternatively, we see that the model trained on set E has much more balanced coefficients than the coefficients of the model trained on set B.

\begin{figure}[t]
\begin{center}
\includegraphics[width=2.5in]{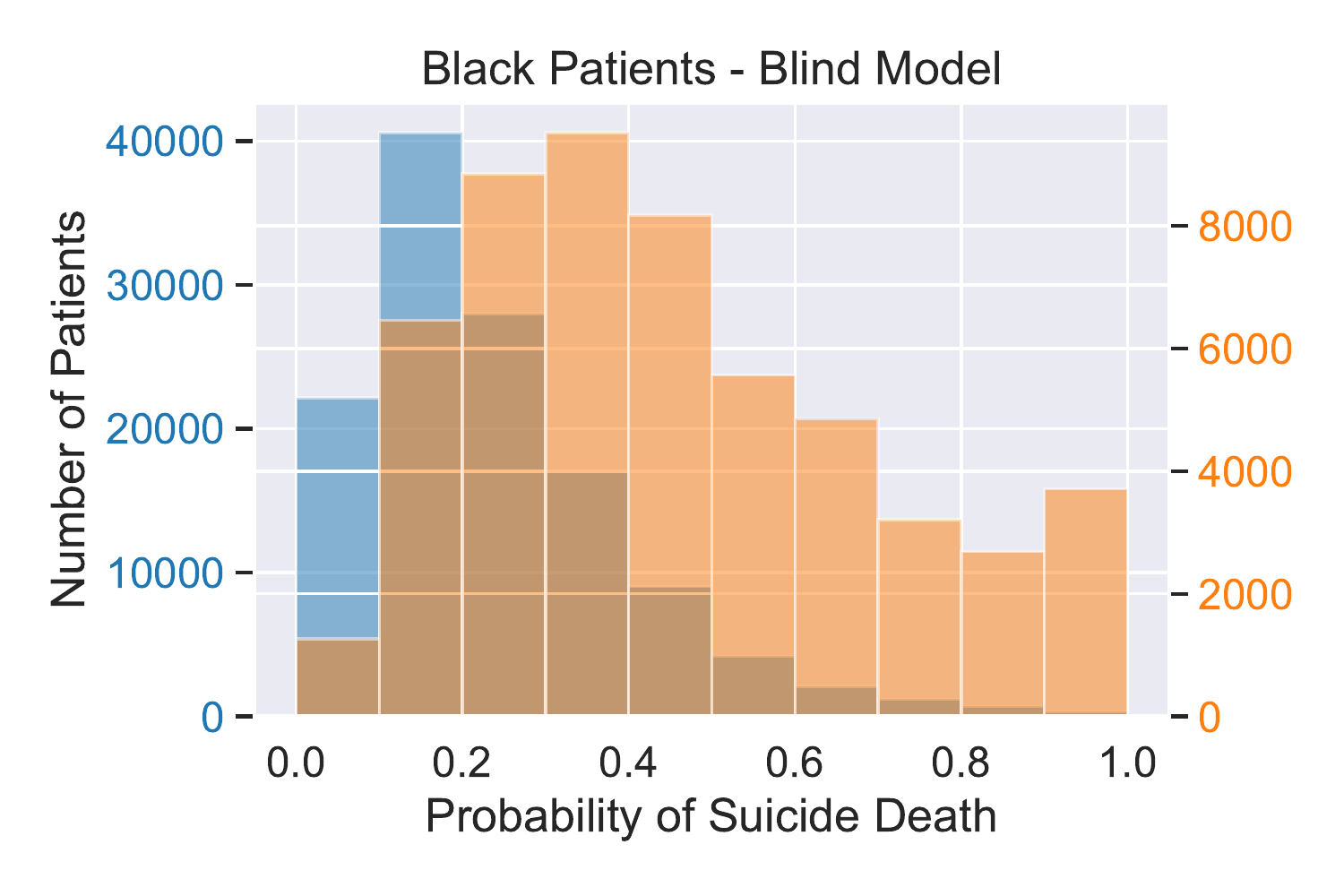}
\includegraphics[width=2.5in]{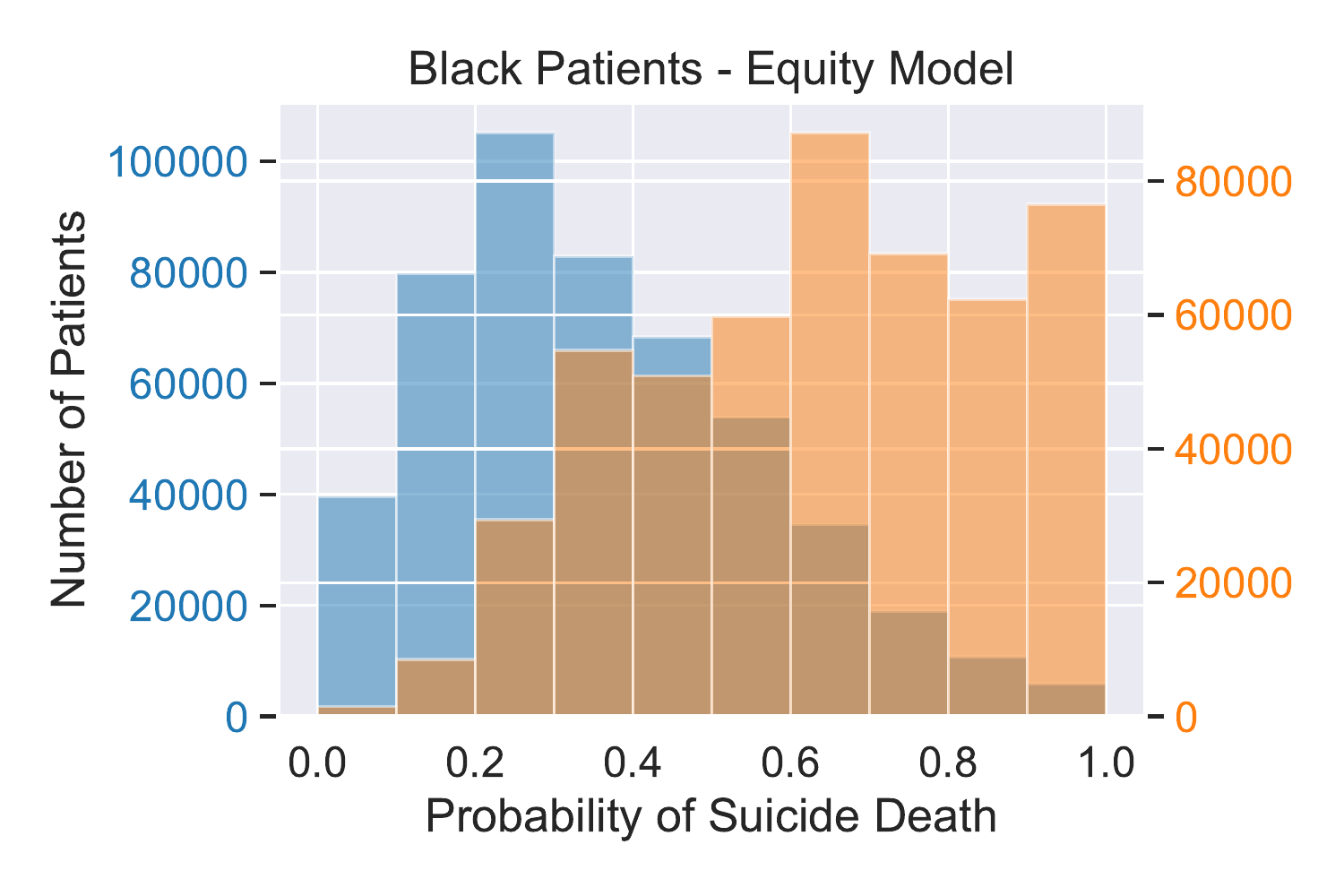} \\
\includegraphics[width=2.5in]{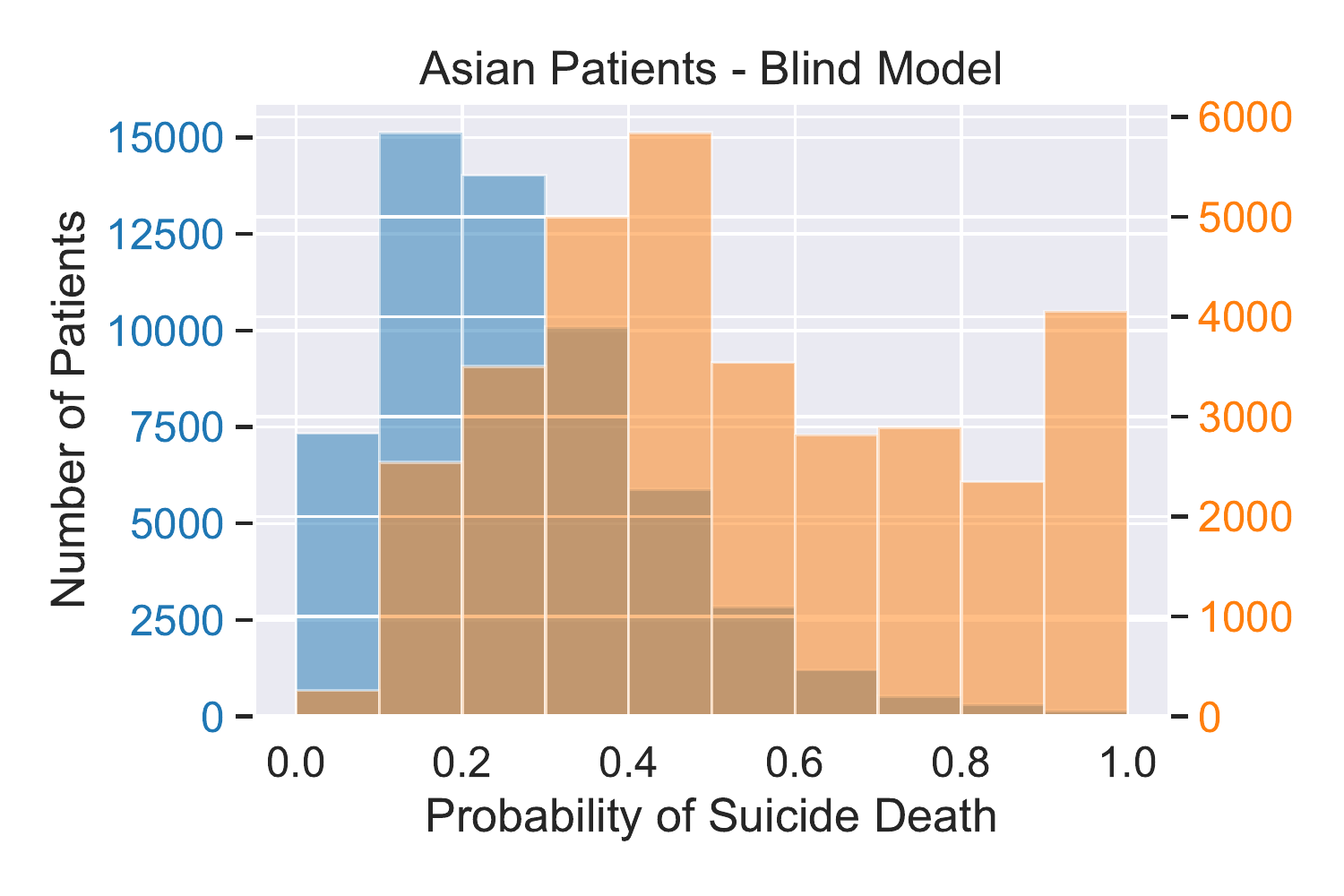} \includegraphics[width=2.5in]{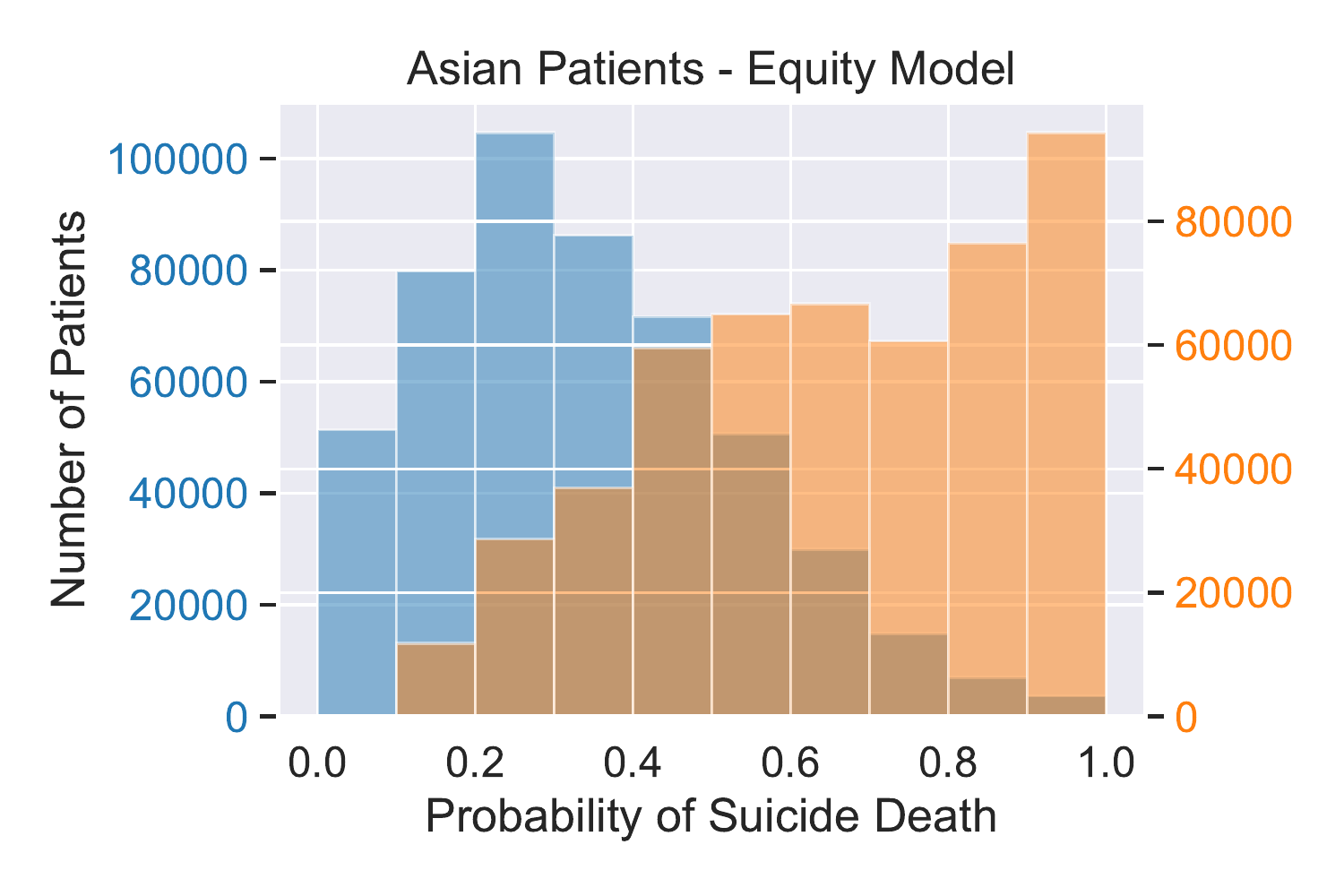}\\
\includegraphics[width=2.5in]{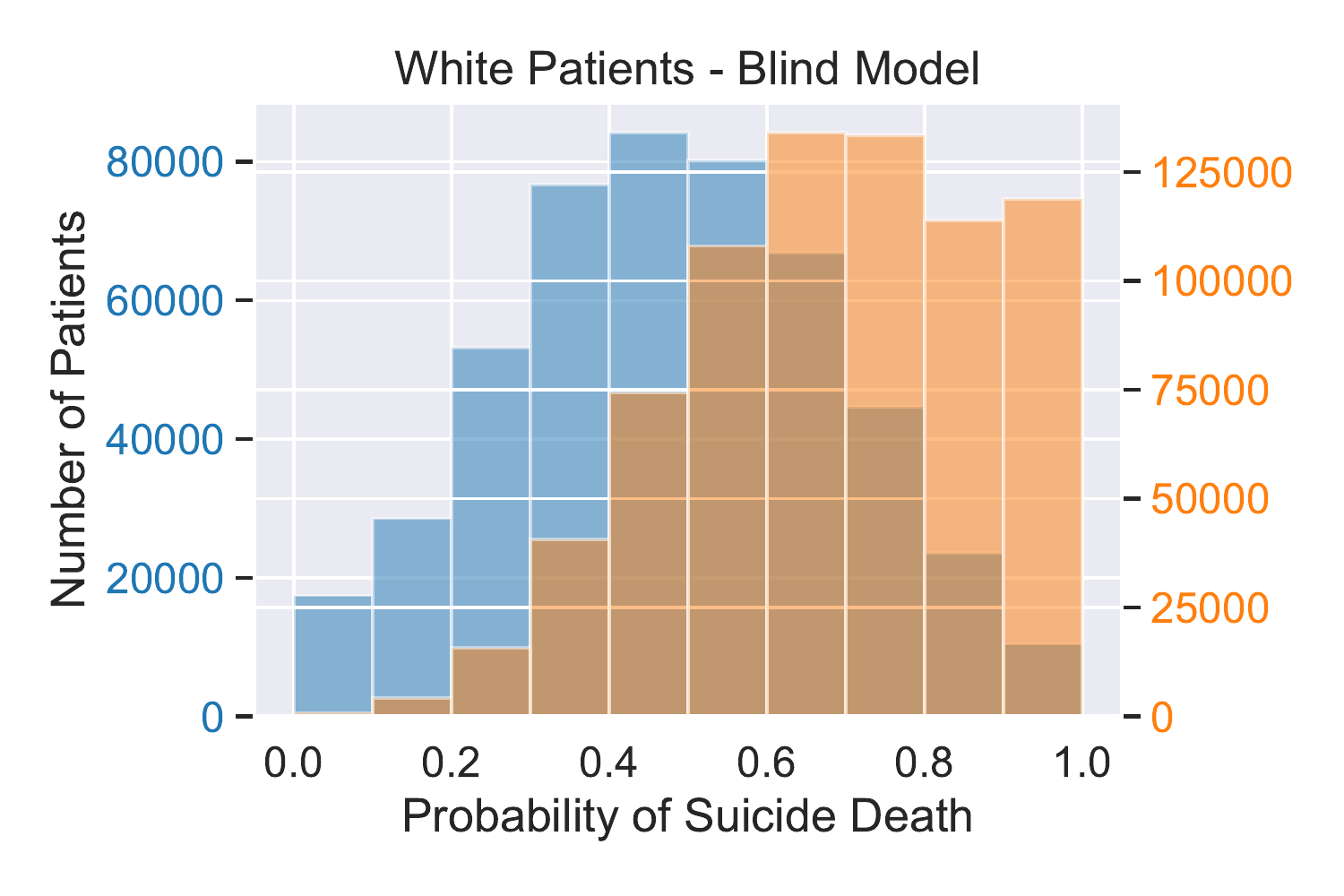} \includegraphics[width=2.5in]{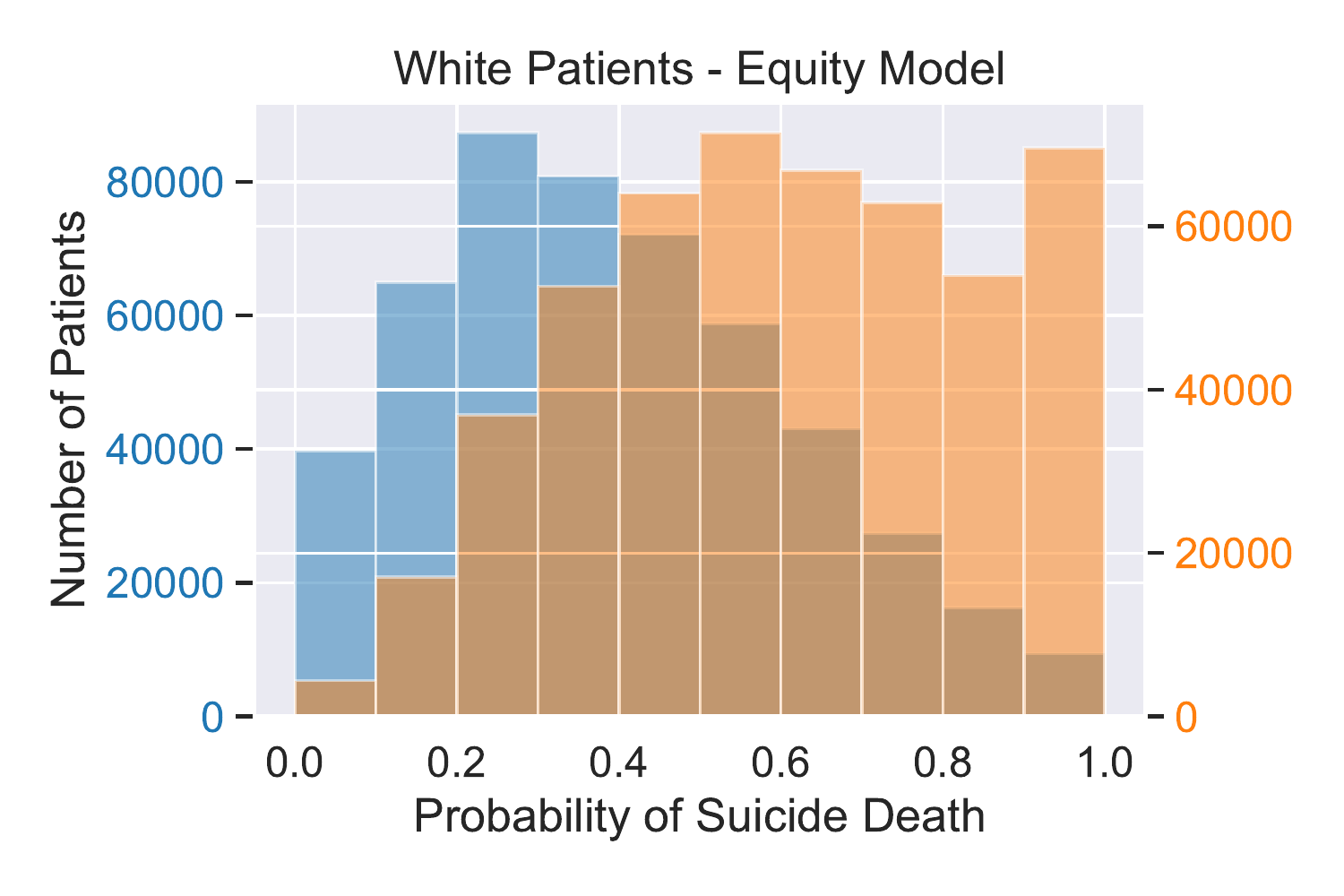} \\
\includegraphics[width=2.5in]{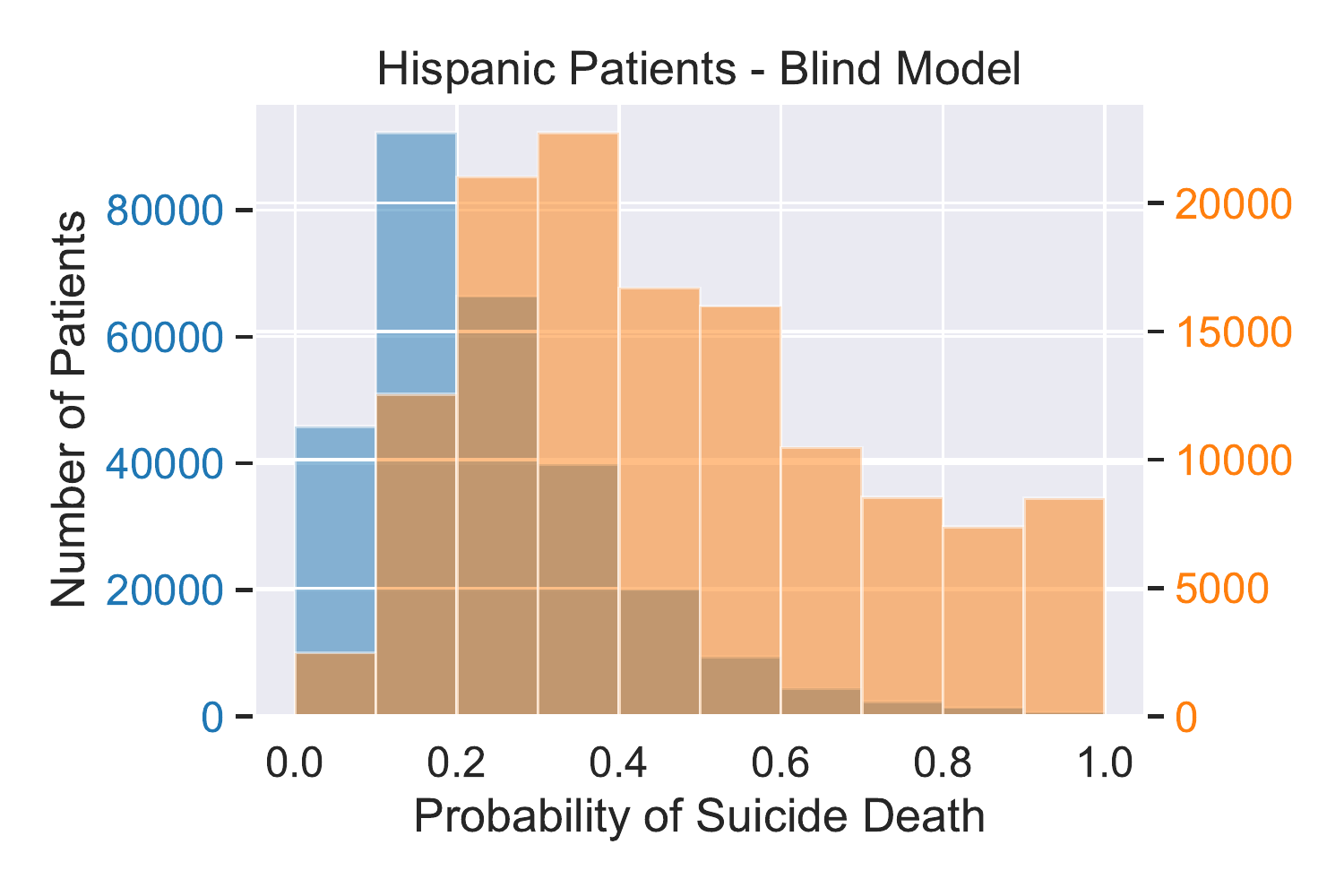}
\includegraphics[width=2.5in]{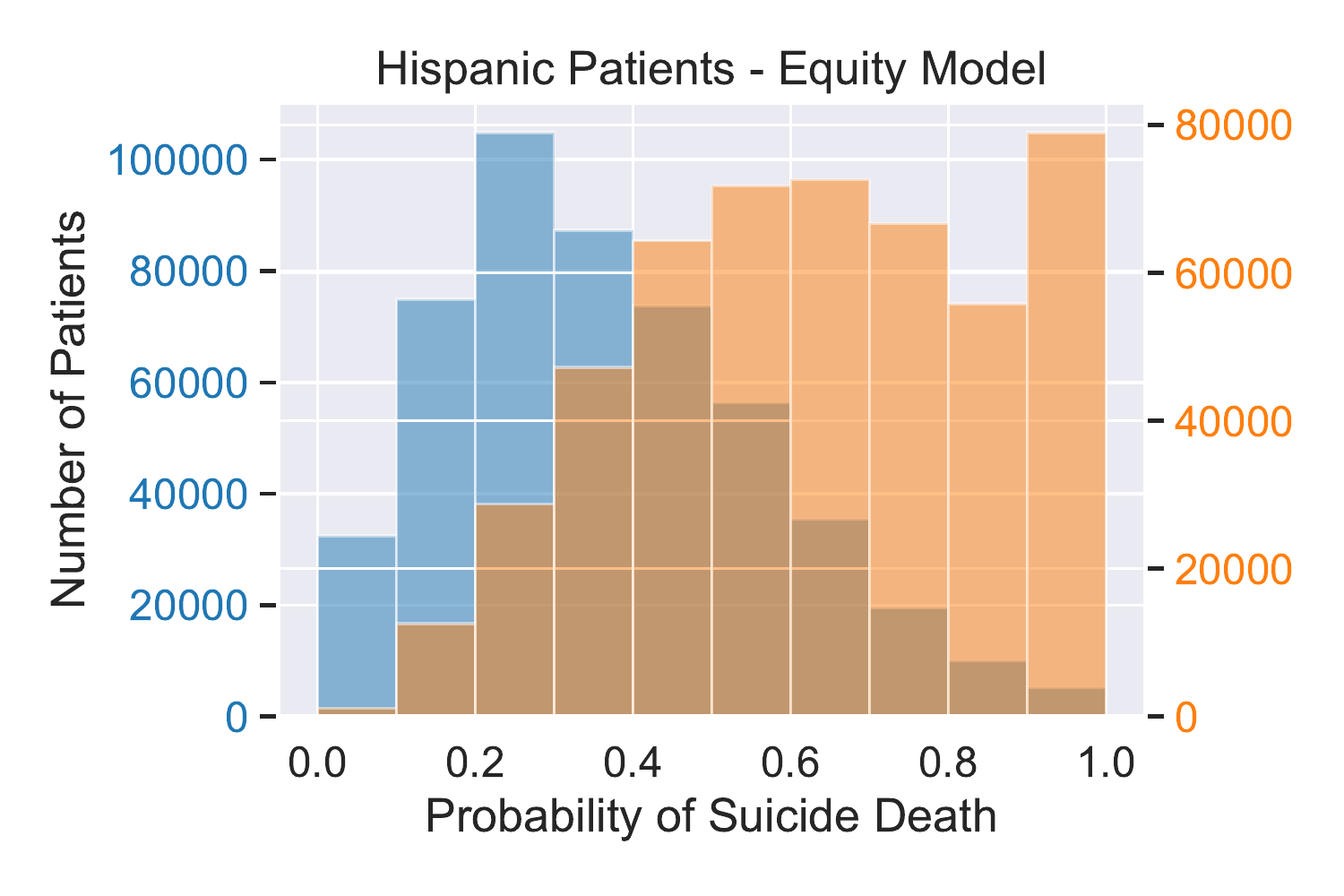}
\end{center}
\caption{\small For our logistic regression models, we examine test set predictions.  In orange, we plot histograms of predicted probability of suicide death for only those patient-visits corresponding to $Y=1$ (suicide death); in blue, we plot a histogram of the same predicted probability for only those patient-visits corresponding to $Y=0$.  The plots on the left (respectively, right) are for the logistic regression model trained on the blind training set $B$ (respectively, equity-directed training set $E$). When we train on the blind training set $B$, the non-White groups' probability of suicide death, regardless of ground truth, are shifted down.  When we retrain the models using the equity-directed bootstrapped data set $E$, we see histograms that are more symmetric and centered; a choice of $\tau=0.5$ is more appropriate here.  One way to achieve the results of the $E$-trained models is to take $B$-trained models and apply group-dependent thresholds $\tau_{a}$.}
\label{fig:bootequityplots}
\end{figure}

We can relate the coefficients in Table \ref{tab:race_coeff} to Figure \ref{fig:bootequityplots}. Looking at the histograms of the probability that a patient record results in suicide death (orange), we see that for all non-White racial/ethnic minority groups, the histograms are shifted left. We also see that for White patients, the records that do not result in suicide death (blue) mimic a normal distribution centered near a probability of $0.5$. This implies that the Blind model is not actually learning an accurate, discriminative model for suicide death but is instead \emph{overrelying} on race/ethnicity as a predictor.  This is simply because in the Blind training set B, of the files corresponding to patients who die by suicide, a majority belong to White patients.

\begin{table}[t]
\begin{center}
\begin{tabular}{lrr}
\hline
          &   Blind Coefficients &   Equity Coefficients \\
\hline
 Black    &                -1.06 &                 -0.31 \\
 Asian    &                -0.69 &                 -0.22 \\
 White    &                 0.12 &                 -0.26 \\
 Hispanic &                -1.02 &                 -0.25 \\
\hline
\end{tabular}
\caption{Logistic regression model coefficients for each racial/ethnic predictor for Blind training data B and Equity training data E.}
\label{tab:race_coeff}
\end{center}
\end{table}

\section{Analysis of the Equity-Directed Bootstrap}
\label{sect:theory}
In this section, we begin the task of providing mathematical justification for the equity-directed bootstrap.  We focus on Na\"ive Bayes and logistic regression, both because our results (see Table \ref{tab:equity}) indicate they are comparable to other methods, and because their form readily admits mathematical analysis.

We first consider how the equity-directed bootstrap affects the na\"ive Bayes classifier.  We partition the predictors via $X = (A,Z)$, where $A$ denotes group membership ({\it e.g.}, racial/ethnic identity) and $Z$ denotes all other predictors.  We begin with Bayes' theorem:
\[
P(Y=y \, | \, A=a, Z=\mathbf{z}) = \frac{ P(A=a,Z=\mathbf{z} \, | \, Y=y) P(Y=y) }{ P(A=a, Z=\mathbf{z}) }.
\]
The left-hand side is the posterior probability of membership in class $y$, given membership in group $a$ and other predictors $\mathbf{z}$.  The numerator on the right-hand side consists of a likelihood times the prior $P(Y=y)$.  We will treat the denominator $P(A = a, Z=\mathbf{z})$ as a normalization constant $\mathcal{C}$.  The assumption of na\"ive Bayes is that the likelihood factors due to independence:
\begin{equation}
\label{eqn:nbassumption}
P(A=a, Z=\mathbf{z} \, | \, Y = y) = \widehat{P}(A = a \, | \, Y = y) \widehat{P}(Z=\mathbf{z} \, | \, Y = y).
\end{equation}
The factorization is typically carried out for all predictors $Z$ as well, but that is unnecessary for our purposes.  Hats denote na\"ive Bayes estimates.  Naturally, as both $A$ and $Y$ are discrete random variables, we estimate the first conditional probability via simple counting,
\[
\widehat{P}(A = a \, | \, Y = y) = \frac{ \sum_{i=1}^N I_{A_i=a} I_{Y_i=y} }{ \sum_{i=1}^N I_{Y_i=y} }.
\]
In the equity-directed bootstrap, the bootstrapped training sets are such that the fraction of instances in class $y$ belonging to group $a$ is simply $1/|\mathcal{A}|$.  In other words, the distribution of $A$ given $Y$ is uniform, $\widehat{P}(A = a \, | \, Y = y) = 1/|\mathcal{A}|$.  As the training set is balanced with respect to the label $Y$, the empirical estimate of $P(Y = y)$ will be $1/2$ in the binary classification case.  Putting things together yields the overall na\"ive Bayes model
\[
\widehat{P}(Y = y \, | \, A = a, Z = \mathbf{z}) = \frac{ \widehat{P}(Z = \mathbf{z} \, | \, Y = y)}{ 2 |\mathcal{A}| \mathcal{C} }.
\]
The right-hand side does not depend on $a$.  For na\"ive Bayes, the equity-directed bootstrap works by constructing training sets that render a patient's racial/ethnic group identity uninformative as to their risk of death by suicide, conditional on all other predictors.

\paragraph{Logistic Regression.}
In the remainder of this section, we consider logistic regression.  Assume we have a training set $\mathcal{D}_n = \{(\bxi_i, {\eta}_i)\}_{i = 1, \ldots, n}$, which consists of independent samples of random variables $(X, Y)$ where $X \in \mathbb{R}^p$ and $Y \in \{0, 1\}$.  As above, let $A$ be the random variable (with finite, discrete sample space $\mathcal{A}$) that denotes group membership, and let $|\mathcal{A}|$ denote the number of elements in $\mathcal{A}$.  In our example problem, $\mathcal{A}$ consists of the four racial/ethnic groups.

We partition $\bxi_i = (\mathbf{a}_i, \mathbf{z}_i)$.  Here $\mathbf{a}_i$ is the one-hot encoding of an independent sample of $A$; to encode membership in group $j$, we set the $j$-th component of $\mathbf{a}_i$ to $1$ and all other components of $\mathbf{a}_i$ to $0$.

The model parameters are $\btheta = (\beta_0, \bbeta)$, the concatenation of a scalar intercept $\beta_0$ with the coefficient vector $\bbeta \in \mathbb{R}^p$.  We then model $Y \, | \, X=\mathbf{x}, \btheta$ as Bernoulli with probability
\begin{equation}
\label{eqn:logisticmodel}
f(\mathbf{x}; \btheta) = (1 + e^{-(\beta_0 + \bbeta^T \mathbf{x} )})^{-1}.
\end{equation}
Let $\widetilde{y}_i = 2 \eta_i - 1$, a conversion of the $\{0,1\}$ labels into $\{-1,1\}$ labels.  With training data $\mathcal{D}_n$ and the form of the model, we derive the negative log likelihood
\begin{equation}
\label{eqn:nll2}
J(\btheta) = -\log L(\btheta) = \sum_{i=1}^n \log (1 + e^{ - \widetilde{y}_i (\beta_0 + \bbeta^T \mathbf{\bxi}_i)} ).
\end{equation}

\paragraph{Analysis of Odds Ratios.} Consider the odds ratio
\begin{equation}
\label{eqn:or}
\rho_{j,k} =  \frac{P(Y=1 \, | \, A=a_j)}{P(Y=0 \, | \, A=a_j)} \bigg/  \frac{P(Y=1 \, | \, A = a_k)}{P(Y=0 \, | \, A = a_k)} .
\end{equation}
If we estimate empirically, using our equity-bootstrapped training set $E$, we find that  $P(Y = y \, | \, A = a_j) = 1/2$ for $y \in \{0,1\}$ and all $a_j \in \mathcal{A}$.  Hence $\rho_{j,k} = 1$ for all $j$ and all $k$.  For the purposes of illustration, consider fitting a logistic regression model where the only predictor is the one-hot encoded group membership vectors $ \mathbf{a} $.  This model can be written
\[
\widehat{P}(Y=1 \, | \, A = \mathbf{a}) = f(\mathbf{a}; \btheta) = (1 + e^{-(\beta_0 + \bbeta^T \mathbf{a})})^{-1}.
\]
We use a hat on the left-hand side to denote the logistic regression model's estimated probability.  Note that $\mathbf{a}$ is a one-hot encoded vector of length $|\mathcal{A}|$.  If we use this simplified logistic regression model to compute the odds ratio above, we obtain
\[
\widehat{\rho}_{j,k} = \exp( \beta_0 + \beta_j ) / \exp( \beta_0 + \beta_k ) = e^{\beta_j - \beta_k}.
\]
Setting $\rho_{j,k} = \widehat{\rho}_{j,k}$, we obtain $e^{\beta_j} = e^{\beta_k}$, implying that $\beta_j = \beta_k$  for all $j$ and all $k$.  In this $Z$-omitted model, the logistic regression coefficients for the group identity variables must all be equal to one another; one can easily verify that this occurs in practice.  Training this hypothetical logistic regression model on equity-bootstrapped data leads to equalized predictions of suicide risk across all groups.

When we include the $Z$ variables, the situation changes in one important way: the model now consists of
\begin{equation}
\label{eqn:logregmod}
\widehat{P}(Y=1 \, | \, A = \mathbf{a}, Z = \mathbf{z}) = f(\mathbf{a}, \mathbf{z}; \btheta) = (1 + e^{-(\beta_0 + \bbeta_a^T \mathbf{a} + \bbeta_z^T \mathbf{z} )})^{-1}.
\end{equation}
Here we have partitioned the coefficients $\bbeta = (\bbeta_a, \bbeta_z)$ to match the dimensions of $\bxi_i = (\mathbf{a}_i, \mathbf{z}_i)$.  We can use this to compute a version of the odds ratio in which each probability is conditioned on $Z=\mathbf{z}$:
\begin{equation}
\label{eqn:lor}
\widetilde{\rho}_{j,k} = \frac{P(Y=1 \, | \, A=a_j, Z = \mathbf{z})}{P(Y=0 \, | \, A=a_j, Z = \mathbf{z})} \bigg/  \frac{P(Y=1 \, | \, A = a_k, Z = \mathbf{z})}{P(Y=0 \, | \, A = a_k, Z = \mathbf{z})} = \exp( \beta_0 + \beta_j ) / \exp( \beta_0 + \beta_k ) = e^{\beta_j - \beta_k}.
\end{equation}
Assume we have a method for drawing samples $\mathbf{z}_i \sim Z|A=a_j$, i.e., $\mathbf{z}_i$ that are independent and identically distributed samples from the random variable $Z$ conditioned on $A=a_j$. Then we can compute a Monte Carlo approximation of the following integral
\begin{multline}
\label{eqn:montecarlo}
\widehat{P}(Y=1 \, | \, A=a_j) \\
= \int_{\mathbf{z}} \widehat{P}(Y=1 \, | \, A=a_j, Z = \mathbf{z}) P(Z=\mathbf{z} \, | \, A=a_j) \, d \mathbf{z} \approx \frac{1}{\nu} \sum_{i=1}^{\nu} (1 + e^{-(\beta_0 + \beta_j + \bbeta_z^T \mathbf{z}_i )})^{-1}.
\end{multline}
In case $Z$ is independent of $A$, then $\mathbf{z}_i \sim Z$.  Using (\ref{eqn:montecarlo}), we can compute all the $\widehat{P}$'s necessary to form
\begin{equation}
\label{eqn:mclor}
\widehat{\rho}_{j,k} = \frac{\widehat{P}(Y=1 \, | \, A=a_j)}{\widehat{P}(Y=0 \, | \, A=a_j)} \bigg/  \frac{\widehat{P}(Y=1 \, | \, A = a_k)}{\widehat{P}(Y=0 \, | \, A = a_k)}.
\end{equation}
Below, in a numerical simulation study, we test our theory that fitting logistic regression models to equity-bootstrapped training data yields either $\widetilde{\rho}_{j,k}$ or $\widehat{\rho}_{j,k}$ odds ratios that approximate the odds ratio $\rho_{j,k} \equiv 1$.

\paragraph{Intercept Adjustment.} In various fields including biostatistics and political science, several authors have considered a problem that is closely related to but different than ours.  Let us think of $Y=1$ and $Y=0$ as cases and controls, respectively.  For a set of samples from the population, we have recorded predictor and response variables for both cases and controls, sufficient information to fit a logistic regression model for $P(Y=1 \, | \, X=\mathbf{x})$.  Suppose we now realize that we have sampled cases and controls at unequal rates.  Prior research tells us that maximum likelihood estimates of the logistic regression model parameters, computed using available samples, can be adjusted to take into account unequal sampling rates.  In fact, as we see below, all that we must adjust is the \emph{intercept} $\beta_0$.  While several authors consider this problem from an asymptotic point of view \citep{KingZeng2001, Wang2020}, here we generalize slightly the derivation of \cite{Rashid2008}.

Let $S \in \{ 0, 1\}$ be a random variable denoting whether an instance has been selected.  In the derivation below, we use $a_j$ (unboldfaced) to denote the $j$-th element of $\mathcal{A}$, the sample space of $A$.  Define $\zeta_{y,j} = P(S = 1 \, | \, Y = 1, A = a_j)$. Via Bayes,
\begin{equation}
\label{eqn:samplingequality}
P(Y=1 \, | \, A=a_j, Z=\mathbf{z}, S=1) 
= \frac{ P(S=1 \, | \, A=a_j, Z=\mathbf{z}, Y=1) P(Y = 1 \, | \, A=a_j, Z=\mathbf{z}) }{ Q },
\end{equation}
where
\begin{multline*}
Q = P(S=1 \, | \, A=a_j, Z=\mathbf{z}, Y=0) P(Y = 0 \, | \, A=a_j, Z=\mathbf{z}) \\
+ P(S=1 \, | \, A=a_j, Z=\mathbf{z}, Y=1) P(Y = 1 \, | \, A=a_j, Z=\mathbf{z}).
\end{multline*}
For $P(Y = 1 \, | \, A=a_j, Z=\mathbf{z})$, we substitute the unadjusted logistic regression model (\ref{eqn:logregmod}) fitted to case and control samples.  We assume that $S$ is independent of $Z$ conditional on $Y$ and $A$.  Then $P(S = 1 \, | \, A=a_j, Z=\mathbf{z}, Y=1) = \zeta_{y,j}$ and we obtain
\begin{equation}
\label{eqn:samplingequality2}
P(Y=1 \, | \, A=a_j, Z=\mathbf{z}, S=1) 
= \frac{ \zeta_{1,j} \widehat{P}(Y = 1 \, | \, A=a_j, Z=\mathbf{z}) }{ \zeta_{0,j} \widehat{P}(Y = 0 \, | \, A=a_j, Z=\mathbf{z}) + \zeta_{1,j} \widehat{P}(Y = 1 \, | \, A=a_j, Z=\mathbf{z}) }.
\end{equation}
We divide numerator and denominator by $\zeta_{0,j} \widehat{P}(Y = 0 \, | \, A=a_j, Z=\mathbf{z})$ to obtain, after straightforward algebra,
\begin{equation}
\label{eqn:samplingequality3}
P(Y=1 \, | \, A=a_j, Z=\mathbf{z}, S=1) = (1 + e^{-(\beta_{0,j}^{\ast} + \beta_j + \bbeta_z^T \mathbf{z} )})^{-1},
\end{equation}
where $\beta_{0,j}^{\ast} = \beta_0 + \log(\zeta_{1,j}/\zeta_{0,j})$.  This is a group-dependent intercept adjustment.  To apply this to our work, we use \emph{empirical estimates of the $Y$-mirrored fraction}---specifically, we set
\begin{equation}
\label{eqn:ourrates}
\zeta_{y,j} =  \sum_{i=1}^n I_{Y_i = 1-y} I_{A_i=a_j} \bigg/ \sum_{i=1}^n I_{A_i=a_j} .
\end{equation}
The idea is to \emph{undo} the sampling that resulted in unequal probabilities of suicide death (in our example above) by racial/ethnic group.  Below, in a simulation study, we explore how well this group-dependent intercept adjustment fares.

\paragraph{Equivalence Between Intercept and Threshold Adjustment.}  Before proceeding, let us remark that forming group-dependent intercepts is, for logistic regression, perfectly equivalent to forming group-dependent thresholds.  With a threshold of $\tau$, we frame our model as
\[
g(\mathbf{x}; \btheta) = (1 + e^{-(\beta_0 + \bbeta^T \mathbf{x})})^{-1} - \tau.
\]
For an input $\mathbf{x}$, we obtain a label of $\{0, 1\}$ depending on the sign of $g(\mathbf{x}; \btheta)$.  Now suppose we wish to mimic the predicted labels of a model with intercept $\breve{\beta}_0$ and threshold $\tau$.  Is there a way to get these labels by keeping our old intercept $\beta_0$ and instead using a new threshold $\breve{\tau}$?  The target model is
\[
\breve{g}(\mathbf{x}; \breve{\btheta}) = (1 + e^{-(\breve{\beta}_0 + \bbeta^T \mathbf{x})})^{-1} - \tau.
\]
We set $\breve{g}(\mathbf{x}; \breve{\btheta}) = 0$ to solve for the decision boundary
\begin{equation}
\label{eqn:brevedb}
e^{-(\breve{\beta}_0 + \bbeta^T \mathbf{x})} = (1-\tau)/\tau.
\end{equation}
We would like to choose a new threshold $\breve{\tau}$ such that $e^{-({\beta}_0 + \bbeta^T \mathbf{x})} = (1-\breve{\tau})/\breve{\tau}$ matches (\ref{eqn:brevedb}) for all $\mathbf{x}$.  This will happen if 
\begin{equation}
\label{eqn:brevetau}
\breve{\tau} = \left( e^{ \breve{\beta}_0 - \beta_0 }(1-\tau)/\tau + 1 \right)^{-1}
\end{equation}
By changing the threshold from $\tau$ to $\breve{\tau}$, we can achieve equivalent predicted labels (for all inputs $\mathbf{x}$) as if we had changed the intercept from $\beta_0$ to $\breve{\beta}_0$.  More precisely, in terms of predicted labels, the $\{ \beta_0, \breve{\tau} \}$ model is equivalent to the $\{ \breve{\beta}_0, \tau\}$ model.

\paragraph{Simulation Test.} We conduct simulations to test whether equity-directed bootstrapping and/or group-dependent intercept adjustment lead to equalized odds ratios for logistic regression models.  We choose a group membership variable $A$ with either $|\mathcal{A}| = 3$ or $|\mathcal{A}| = 10$ possible values.  We then augment $A$ with a $p$-dimensional vector of additional predictors; here $p=20$.  In the \emph{discrete} version of the simulation, we take $Z \in \{0, 1\}^p$ where each $Z_i \sim \text{Ber}(0.5)$.  In the \emph{continuous} version of the simulation, we take $Z \in \mathbb{R}^p$ to be multivariate normal with prescribed mean vector and covariance matrix.  We experiment with four different versions of $Z$ corresponding to \emph{zero mean} vs randomly sampled \emph{non-zero mean} (with each element itself sampled from a standard normal), and identity covariance (\emph{uncorrelated}) vs randomly sampled covariance (\emph{correlated}).  To form random covariance matrices $\Sigma$, we take a $p \times p$ matrix $\Phi$ of samples from a standard normal and then set $\Sigma = \Phi^T \Phi$ where $^T$ denotes transpose.

For each choice of $A$ and $Z$, we sample $n=50000$ rows to form an overall design matrix $X$ of size $n \times (1 + |\mathcal{A}| + p)$.  The first column of $X$ is all $1$'s, equivalent to including an intercept.  Then, using a prescribed $\btheta$ vector, we sample $Y \sim \text{Ber}(\sigma(X \btheta)) \in \{0,1\}$ to generate $n$ labels.  Note that we always take the intercept $\beta_0$ and final $p$ elements of $\btheta$ to be random, sampled from $\text{Unif}(-0.1,0.1)$.  In the case where $|\mathcal{A}|=3$, we set $\beta_1 = -0.5$, $\beta_2 = 0.2$, and $\beta_3 = 1.0$, leading to a scenario where the odds ratios are \emph{known to be unequal before we apply any equity-directed methods}.  In the case where $|\mathcal{A}|=10$, we sample $\{\beta_4, \ldots, \beta_{10}\}$ from a $\text{Unif}(-0.1,0.1)$ distribution.

With this sampled data, we compute empirical odds ratios---a simple counting estimate of (\ref{eqn:or}) for each $j,k$---and also fit a logistic regression model (LR).  We then apply equity-directed bootstrapping (sampling a total of $2 |\mathcal{A}| \cdot 800$ elements), recompute empirical odds ratios, and refit a logistic regression model (LR').  For the logistic regression models, we compute odds ratios using both conditional (\ref{eqn:lor}) and Monte Carlo (\ref{eqn:mclor}) estimates.   We also return to the original LR model, apply the group-dependent intercept adjustment, and record the subsequent odds ratios computed via (\ref{eqn:mclor}).

We carry out the above procedure $100$ times.  Our theory above indicates that equity-directed methods should make the matrix of odds ratios equal to one, i.e., $\rho_{j,k} \equiv 1$.  Hence we measure the \emph{mean absolute deviation from one}, for each entry of each odds ratio matrix that we compute.  We present our results in Table \ref{tab:simulation}.  Overall, our results are consistent with the theory presented above: both the equity-directed bootstrap and group-dependent intercept adjustment result in logistic regression models that yield odds ratios much closer to $1$ than in  raw, unadjusted data/models.  Note that the intercept adjustment method performs no worse than the equity-directed bootstrap; both intercept and threshold adjustment may be preferable for problems in which \emph{retraining} a logistic regression model is prohibitively expensive.

Interestingly, we see that when we introduce correlation in the non-group predictor matrix, the performance of all equity-directed methods degrades slightly.  We conjecture that this arises due to accidental correlation between non-group ($Z$) and group ($A$) variables, thus violating the assumptions of independence or conditional independence made above.

In Table \ref{tab:race_coeff}, we record the actual values of the $\beta_i$ logistic regression coefficients before and after applying the equity-directed bootstrap.  Note that the mean absolute deviation from one for the logistic regression model with blind training set $B$ is $0.5666$, whereas for the model with equity-directed training set $E$ it is $0.0350$.  The non-zero value corresponds to small differences in  coefficients corresponding to racial/ethnic group identity.  We hypothesize that these differences may be due to correlations between group and non-group predictors.

\begin{table}[t]
\begin{center} \small
\begin{tabular}{l|l|rrr|rrrr} 
& \text{$Z$ (non-group predictor) distribution} & \multicolumn{3}{c|}{Original Data/Models} & \multicolumn{4}{c}{Equity-Adjusted Data/Models}  \\ 
& & EOR & LOR & MCLOR & EOR & LOR & MCLOR & INTADJ \\ \hline
\parbox[t]{5mm}{\multirow{5}{*}{\rotatebox[origin=c]{90}{$|\mathcal{A}|=3$}}} & Discrete & 0.8343 & 0.8386 & 0.8336 & 0.0000 & 0.0045 & 0.0045 & 0.0023 \\
& Continuous, zero mean, uncorrelated & 0.8242 & 0.8427 & 0.8240 & 0.0000 & 0.0107 & 0.0105 & 0.0100 \\
& Continuous, zero mean, correlated & 0.6239 & 0.8433 & 0.6250 & 0.0000 & 0.1340 & 0.1039 & 0.1050 \\
& Continuous, random mean, uncorrelated & 0.8231 & 0.8417 & 0.8229 & 0.0000 & 0.0121 & 0.0118 & 0.0100 \\
& Continuous, random mean, correlated & 0.6190 & 0.8418 & 0.6196 & 0.0000 & 0.1375 & 0.1064 & 0.1064 \\ \hline
\parbox[t]{5mm}{\multirow{5}{*}{\rotatebox[origin=c]{90}{$|\mathcal{A}|=10$}}} & Discrete & 0.3714 & 0.3729 & 0.3708 & 0.0000 & 0.0050 & 0.0050 & 0.0024 \\
& Continuous, zero mean, uncorrelated &0.3673 & 0.3740 & 0.3667 & 0.0000 & 0.0101 & 0.0099 & 0.0065 \\
& Continuous, zero mean, correlated & 0.2911 & 0.3734 & 0.2898 & 0.0000 & 0.0732 & 0.0579 & 0.0519 \\
& Continuous, random mean, uncorrelated &0.3650 & 0.3716 & 0.3644 & 0.0000 & 0.0099 & 0.0097 & 0.0065 \\
& Continuous, random mean, correlated &0.2881 & 0.3738 & 0.2876 & 0.0000 & 0.0753 & 0.0591 & 0.0535
\end{tabular}
\end{center}
\caption{Simulation results showing mean absolute deviation of odds ratios from one.  We see that both the equity-directed bootstrap and group-dependent intercept adjustment yield odds ratio matrices that are close to $1$; this is consistent with our theory in Section \ref{sect:theory}.  Here EOR stands for an empirical estimate of the odds ratios (\ref{eqn:or}), LOR stands for the conditional logistic regression odds ratios (\ref{eqn:lor}), MCLOR stands for the Monte Carlo estimate of the logistic regression odds ratios (\ref{eqn:mclor}), and INTADJ stands for the Monte Carlo estimate of the odds ratios of the logistic regression model with group-dependent intercept adjustment.}
\label{tab:simulation}
\end{table}

\paragraph{Asymptotic Analysis of the Optimal Solution.} The gradient and Hessian of (\ref{eqn:nll2}) can be derived straightforwardly and written in the following form \citep{murphy2012machine}:
\begin{align*}
\mathbf{g}(\btheta) &= \nabla_{\btheta} [-\log L(\btheta)] = \bXi^T (\bmu - \bleta)  \\\mathbf{H}(\btheta) &= \nabla_{\btheta} \nabla_{\btheta}[-\log L(\btheta)] = \bXi^T \operatorname{diag}[ \bmu (1 - \bmu) ] \bXi.
\end{align*}
Here $\bXi$ is the $n \times (p+1)$ matrix whose $i$-th row is $(1,\bxi_i)$, and $\bmu$ is the vector whose $i$-th component $\mu_i$ is the probabilistic prediction on the $i$-th training instance, i.e., $\mu_i = f(\bxi_i; \btheta)$ with $f$ as in (\ref{eqn:logisticmodel}).  For finite $\mathbf{x}$ and $\btheta$, we see that $\mu_i \in (0,1)$.  Thus for any $\bXi$, the Hessian $\mathbf{H}$ will be positive semi-definite, implying that $J$ is convex.  Hence the set of global minimizers of $J$ is not empty; let $\btheta^{\ast}$ denote any member of this set.  When we \emph{train} the logistic regression model on data $\mathcal{D}_n$, we apply an optimization algorithm ({\it e.g.}, Newton's method) to compute an approximation to such a $\btheta^{\ast}$.
To further analyze equity-directed bootstrapping, we analyze the properties that $\btheta^{\ast}$ must satisfy.  To be a minimizer, it must satisfy $\mathbf{g}(\btheta^{\ast}) = \mathbf{0}$:
\begin{equation*}
\sum_{i=1}^n (\mu_i(\btheta^{\ast}) - \eta_i) = 0, \text{ and } \sum_{i=1}^n (\mu_i(\btheta^{\ast}) - \eta_i) \xi_{i,j} = 0 \ \text{ for all } j.
\end{equation*}
Restricting attention to $j = 1, \ldots, |\mathcal{A}|$, we have
\begin{equation}
\label{eqn:almostthere}
\sum_{i=1}^n (\mu_i(\btheta^{\ast}) - \eta_i) a_{i,j} = 0 \quad \Longrightarrow  \sum_{i \text{ s.t. } a_{i,j}=1} \mu_i(\btheta^{\ast}) = \sum_{i \text{ s.t. } a_{i,j}=1} \eta_i,
\end{equation}
as $a_{i,j} = 1$ if and only if row $i$ corresponds to a patient in group $j$.
The right-hand side is the number of rows corresponding to patients in group $j$ who have died by suicide.  If our training set $\mathcal{D}_n$ is the equity-directed bootstrap set $E$ described above, the right-hand side is precisely $M$.  Note also that we must have $n = 2 | \mathcal{A}| M$, where $n$ is the total number of rows in the training set.  Partitioning $\btheta^{\ast} = (\beta_0^{\ast}, \bbeta_a^\ast, \bbeta_z^\ast)$ and using the fact that $\mathbf{a}_i$ is a unit vector, (\ref{eqn:almostthere}) becomes
\[
\frac{|\mathcal{A}|}{n} \sum_{i \text{ s.t. } a_{i,j}=1} (1 + \exp[-(\beta_0^{\ast} + \beta^{\ast}_{a,j} + (\bbeta_z^\ast)^T \mathbf{z}_i)])^{-1}
 = \frac{1}{2}.
\]
For the equity-directed training set $E$, there are precisely $n/|\mathcal{A}|$ terms in the sum, and hence the left-hand side is a sample average.  Under mild assumptions on the random variables $X$, this will converge via the law of large numbers to a constraint on expected values
\[
E_{Z  |  A = a_j} \left[ (1 + \exp[-(\beta_0^{\ast} + \beta^{\ast}_{a,j} + (\bbeta_z^\ast)^T Z)])^{-1} \right] = \frac{1}{2}.
\]
Here the expected value is over the distribution of $Z$ given that the group identity $A$ is $a_j$, the $j$-th element of $\mathcal{A}$.  Let us invoke an assumption that is analogous to but stronger than (\ref{eqn:nbassumption})---namely, assume that the predictors $Z$ ({\it i.e.},the predictors \emph{other} than group identity $A$) are independent of $A$.  Then the left-hand side becomes
\[
E_{Z} \left[ (1 + \exp[-(\beta_0^{\ast} + \beta^{\ast}_{a,j} + (\bbeta_z^\ast)^T Z)])^{-1} \right] = \frac{1}{2}.
\]
Now the left-hand side depends on $j$ only through $\beta^{\ast}_{a,j}$, while the right-hand side does not depend on $j$ at all.  As the sigmoid function $\sigma(z) = (1+\exp(-z))^{-1}$ is strictly monotonic, the only way this can hold for all $j$ simultaneously is if $\beta^{\ast}_{a,j}$ is constant with respect to $j$.

In the $n \to \infty$ large sample limit, and assuming independence of $A$ and $Z$, training with the equity-directed bootstrap yields group identity coefficients that \emph{are equal to each other}.  This matches our reasoning and results with odds ratios above.

%, we can write (\ref{eqn:almostthere}) as
%\[
%\sum_{i \text{ s.t. } a_{i,j}=1} (1 + \exp[-(\beta_0^{\ast} + (\bbeta_a^\ast)^T \mathbf{a}_i + (\bbeta_z^\ast)^T \mathbf{z}_i)])^{-1}
% = \frac{n}{2 |\mathcal{A}|}.
%\]
%Inside the sum, we know that 

\paragraph{Relationship Between Resampling and Weighting.}
Starting from the equity-directed bootstrap, we derive an equity-directed \emph{reweighting} of the loss function.  We carry out this derivation for the logistic regression loss, but in principle this can be applied to other methods that involve empirical risk minimization, {\it e.g.}, neural networks.

Let $N$ be the total number of instances in the original, raw training set.  Let $J(\btheta)$ denote the negative log likelihood (\ref{eqn:nll2}).  We rewrite this by separating out the class and group labels:
\[
J(\btheta) = \sum_{i=1}^N \sum_{a \in \mathcal{A}} \sum_{y \in \{-1,1\}} I_{A_i = a} I_{y_i = y} \log(1 + e^{-y \btheta^T \bxi_i} ) = \sum_{i=1}^{n_a^y} \log(1 + e^{-y \btheta^T \bxi_i^{a,y}} ),
\]
where $n_a^y$ denotes $\sum_{i=1}^N I_{A_i = a} I_{y_i = y}$, the number of instances with group $a$ and class $y$; $\bXi^{a,y}$ is the restriction of the predictor matrix to only those rows corresponding to class $y$ and group $a$; and $\bxi_i$ is the $i$-th row of $\bXi^{a,y}$.  Let $M$ denote the desired number of instances in each equity-directed bootstrap subgroup; in the above example, we set $M=500000$.  Let $Z^{a,y}$ denote a random vector of length $n_a^y$ whose entries are nonnegative integers and whose sum is constrained to be exactly $M$.  It is sufficient for $Z^{a,y}$ to have a Dirichlet-multinomial distribution with parameters $M$ and $\alpha_i = 1$ for $i = 1, \ldots, n_a^y$.  Then we will have $E[Z^{a,y}_i] = M/n_a^y$.  Consider 
\begin{equation}
\label{eqn:equityloss}
J_1(\btheta) = \sum_{i=1}^{n_a^y} Z^{a,y}_i \log(1 + e^{-y \btheta^T \bxi_i^{a,y}} )
\end{equation}
A realization of the stochastic objective $J_1(\btheta)$, for a particular sample of $Z_a^y$, is equivalent to the equity-directed bootstrap.  If we take the \emph{expected value} of the stochastic objective $J_1$, we obtain
\begin{equation}
\label{eqn:weightedloss}
J_2(\btheta) = \sum_{i=1}^{n_a^y} \frac{M}{n_a^y} \log(1 + e^{-y \btheta^T \bxi_i^{a,y}} )
\end{equation}
This is a deterministic objective function with label and group-dependent \emph{weights} that serve the same purpose as the equity-directed bootstrap.  We could refer to (\ref{eqn:weightedloss}) as an equity-weighted loss function.  In practice, we do not pursue minimization of (\ref{eqn:weightedloss}) because it still involves sums over $N$ instances.  Any realization of (\ref{eqn:equityloss}), on the other hand, will have precisely $C M$ terms in the sum with $C = 2 |\mathcal{A}|$.  

\section{Discussion}
\label{sect:discussion}

In this work, we have proposed an equity-directed bootstrap for problems in which one desires to achieve the equalized odds criterion across subgroups of the population.  We have demonstrated that the bootstrap helps to bring models closer to equalized odds on real data (the suicide prediction problem).  We have taken a first pass at understanding how and why the bootstrap works, by examining odds ratios in the context of logistic regression and na\"ive Bayes.  We find that simulations are consistent with our theory.

The present work suggests tasks and questions for future work.  First, our results on the suicide prediction problem indicate that, as predictive modeling with health care data sets becomes more common, researchers should interrogate the algorithmic fairness of such models.  We do not believe this is common practice yet in the literature.  Second, equal odds is one of many algorithmic fairness criteria.  Are there versions of the bootstrap that can help achieve algorithmic equity in other ways?  Finally, a limitation of our results is their reliance on conditional independence assumptions.  In future work, we seek extensions of the bootstrap in which this assumption can be relaxed.

\section*{Acknowledgments}
Funding for this project was provided by the University of California Firearm Violence Research Center (to SGM and HSB), National Institute of Mental Health grant R15 MH113108-01 (to SGM), NSF grant DGE-1633722 (through which MER was funded through an NRT graduate fellowship), and NSF grant ACI-1429783 (for computational time on the MERCED cluster).

\subsection*{Data Availability Statement}
We cannot share Administrative Patient Records; all code used to generate simulation results is available upon request.

\subsection*{Financial disclosure}

None reported.

\subsection*{Conflict of interest}

The authors declare no potential conflict of interests.

%\bibliographystyle{abbrvnat}
%\bibliography{draft}

\end{document}